\documentclass{article}

\usepackage{arxiv}

\usepackage[utf8]{inputenc} 
\usepackage[T1]{fontenc}    
\usepackage{hyperref}       
\hypersetup{
    colorlinks=true,
    linkcolor=blue,
    citecolor=blue,
    filecolor=magenta,      
    urlcolor=black,
}
\usepackage{url}            
\usepackage{booktabs}       
\usepackage{amsfonts}       
\usepackage{nicefrac}       
\usepackage{microtype}      
\usepackage{lipsum}
\usepackage[numbers]{natbib}
\usepackage{array}

\title{ELECTRAMed: a new pre-trained language representation model for biomedical NLP}

\author{
Giacomo Miolo\thanks{Corresponding authors.}\,\,\,, Giulio Mantoan$^\ast$, Carlotta Orsenigo \\
Department of Management, Economics and Industrial Engineering \\
Politecnico di Milano, Italy \\
\texttt{\{giacomo.miolo,giulio.mantoan,orsenigo\}@mip.polimi.it} \\
}

\begin{document}
\maketitle

\begin{abstract}
The overwhelming amount of biomedical scientific texts calls for the development of effective language models able to tackle a wide range of biomedical natural language processing (NLP) tasks. The most recent dominant approaches are domain-specific models, initialized with general-domain textual data and then trained on a variety of scientific corpora. However, it has been observed that for specialized domains in which large corpora exist, training a model from scratch with just in-domain knowledge may yield better results. Moreover, the increasing focus on the compute costs for pre-training recently led to the design of more efficient architectures, such as ELECTRA. In this paper, we propose a pre-trained domain-specific language model, called ELECTRAMed, suited for the biomedical field. The novel approach inherits the learning framework of the general-domain ELECTRA architecture, as well as its computational advantages. Experiments performed on benchmark datasets for several biomedical NLP tasks support the usefulness of ELECTRAMed, which sets the novel state-of-the-art result on the BC5CDR corpus for named entity recognition, and provides the best outcome in 2 over the 5 runs of the 7th BioASQ-factoid Challange for the question answering task.\\
\end{abstract}


\section{Introduction}
The immense body of biomedical scientific texts, which steadily grows at an exponential rate, makes it imperative to develop effective machine learning methods able to automatically extract the rich knowledge therein contained, that can be used to address several biomedical natural language processing (NLP) tasks.

Prominent NLP advancements in recent years have been mostly driven by the use of deep neural models, which require large corpora of annotated training data. 
Compared to the general domain, however, the collection of such data in the biomedical field is difficult and expensive, since it  necessarily involves domain experts for accurate data labelling.  
For this reason, semi-supervised pre-trained language models, such as ELMo \citep{Peters18} and BERT \citep{Devlin19}, were developed and successfully applied in a wide range of NLP tasks.

ELMo and BERT leverage contextualized word embeddings for which the representation of a word depends on the context where it is used and, therefore, is a function of the entire input sequence. 
ELMo exploits a deep bidirectional language model pre-trained on a large corpus of texts. It was proposed to address both the syntactic and semantic complexities and ambiguities of words and has been proven to achieve notable  results in a variety of NLP problems \citep{Peters18}.
BERT \citep{Devlin19} resorts to the transformer architecture to pre-train bidirectional language representations, instead of relying on recurrent neural networks.
By embracing attention mechanisms, BERT showed to distinguish the sense of words at a very fine level and to grasp many of their syntactic and semantic properties.
This ability made BERT the state-of-the-art contextualized word representation model for the most challenging natural language understanding problems.

While ELMo and BERT architectures pre-trained on general-domain corpora are well-established top performers for general NLP tasks, they might yield poor results in case of scientific or specific domains, since the corpora used for pre-training, such as news articles and Wikipedia \citep{Belinkov19}, might not include the same terminology adopted in the in-domain tasks.
For specialized contexts past studies showed that general-domain language models can largely benefit from the use of in-domain textual data \citep{Peng19}.
As a consequence, recent models for biomedical NLP relied on adapted versions of general-domain approaches. 
Among these, two of the most noteworthy and successful examples are represented by BioBERT \citep{Lee2020} and BlueBERT \citep{Peng19}, which are domain-specific language models initialized with the general-domain BERT, and then pre-trained on a wide range of biomedical and scientific corpora. 
In principle, these last methods rely on the assumption that initializing the pre-training from general-domain models might improve the overall performance for domain-specific purposes.
However, it has been recently observed that for domains in which large corpora exist, like the biomedical field, pre-training language models from scratch yields better results than feeding the pre-training phase with general-domain knowledge \citep{Gu20}.

To obtain contextualized word embeddings, BERT pre-training is based on masked language modelling (MLM) which aims at predicting a small  random subset of masked input tokens, considering only the token context.
This approach allows the model to learn bidirectional representations.
A different input corruption procedure has been recently proposed, in which instead of masking, and therefore losing, some of the input tokens, these are replaced with plausible alternatives produced by a small generator network. 
By learning from all the input tokens this novel approach, called ELECTRA (Efficiently Learning an Encoder that Classifies Token Replacements Accurately), is computationally much more efficient than BERT, and has been shown to outperform the latter in several tasks \citep{Clark20}.

Inspired by previous research achievements which showed the effectiveness of language models built on domain-specific knowledge, and recognizing the importance of resorting to efficiently pre-trained methods while preserving downstream performances, in this study we describe a new ELECTRA-based model, called ELECTRAMed, suited for the biomedical domain.  

In particular, the main contributions of the present work are as follows:
\begin{itemize}
    \item We propose a novel ELECTRA-based language representation model (ELECTRAMed) pre-trained on biomedical corpora. To the best of our knowledge, this is the first ELECTRA-based model specifically developed for the biomedical domain, and the present study is the first which applies, to this domain, a transformer architecture different from BERT.
    \item We tested ELECTRAMed on several biomedical benchmark NLP tasks. The results achieved empirically show the effectiveness of the proposed approach, which performed at par, and sometimes better, than state-of-the-art models while leveraging the reduced computational effort required by ELECTRA-based architectures.
    \item We make publicly available the pre-processed datasets used in our study as well as the pre-trained weights of ELECTRAMed and the source code for fine-tuning the model\setcounter{footnote}{0}\footnote{https://github.com/gmpoli/electramed}.
\end{itemize}

\section{Materials and methods}
As it is generally the case with pre-trained methods for language representation, the development of the new model encompassed two main phases represented, respectively, by pre-training and fine-tuning, as illustrated in the following sections.
Since the proposed approach shares the same architecture of ELECTRA, a brief description of the latter is also required.

\subsection{ELECTRA}
ELECTRA (Efficiently Learning an Encoder that Classifies Token Replacements Accurately) is a pre-trained language model recently proposed to overcome the computational drawback of approaches based on masked language modelling (MLM), like BERT, where the input is first corrupted by masking some tokens in the sentence and a network is then trained to recover the original identity of the corrupted tokens. 
Since the network learns from a small percentage of masked-out tokens (typically 15\%), these techniques usually require a huge amount of computations to be effective \citep{Clark20}.

To address this inefficiency, ELECTRA implements an alternative pre-training procedure called Replaced Token Detection (RTD), in which a network is trained to distinguish real input tokens from synthetic but plausible replacements.
Specifically, the model consists of two neural networks which are pre-trained jointly. 
The first is a generator that performs MLM by providing tokens substitutes, and that then learns to predict the original token from the masked form. The second is a discriminator which is trained to detect the synthetic tokens, i.e. to distinguish real tokens from those replaced by the generator.
After pre-training the generator is dropped and the discriminator is fine-tuned on labeled data for downstream tasks.

Compared to MLM, the innovative pre-training procedure embedded in ELECTRA is more computationally efficient since the discriminator is required to predict the class, real vs. replaced, of each of the input tokens, thereby learning from the entire input sequence instead of a small portion of it.
This feature lets ELECTRA compete with state-of-the-arts models in several NLP problems while reducing dramatically the computational effort needed for pre-training, as described in the seminal paper \citep{Clark20}. Moreover, by resorting to plausible tokens alternatives for masking, RTD helps in mitigating a disadvantage of the traditional MLM approach, which introduces a potential mismatch in the learning framework between pre-training and fine-tuning due to the use, for input corruption, of the same conventional [MASK] token which does not appear in the fine-tuning phase.
Its computational efficiency, together with the promising results achieved for different NLP tasks, motivated the adoption of ELECTRA in the present study.

\subsection{ELECTRAMed pre-training}
Differently from other prominenent pre-trained models, such as BioBERT, ELECTRAMed wasn't initialized with the weights from ELECTRA.
Indeed, pre-training was performed entirely on a biomedical domain corpus.
The corpus at hand was published by \citep{Peng19} and consists of 28,714,373 PubMed abstracts (uncompressed, approximately 26GB), representing the whole amount of abstracts published on the free digital repository until September 2018.
In more detail, the corpus contains $\sim$181M sentences and $\sim$4B words and was subject to the following common NLP pre-processing steps applied by the  proponents of the dataset:
\begin{itemize}
    \item Text lowercasing;
    \item Special characters (\textbackslash x00-\textbackslash x7F) removal;
    \item Text tokenization using NLTK Treebank tokenizer \footnote{http://www.nltk.org/}.
\end{itemize}

In any NLP task, vocabularies are needed to encode tokens with numbers, and are generally built so to contain the most frequent words or subword units.
In the present study we made use of SciVocab, a WordPiece vocabulary proposed by \citep{Beltagy19} and built on a scientific text corpus by means of the SentencePiece library \footnote{https://github.com/google/sentencepiece}. 
Compared to the more commonly used vocabulary released with BERT, SciVocab is characterized by almost the same size ($\sim$30k) and 42\% of tokens overlap, thereby showing a substantial difference in the words used frequently in scientific texts with respect to the general domain case.
Notice that, the use of a more scientific-oriented vocabulary should reduce the incidence of out-of-vocabulary (OOV) tokens and, therefore, the loss of information in the text encoding phase.
The hyperparameters applied for pre-training ELECTRAMed on the corpus mentioned above are reported in Table~\ref{pretraining_hp}.
These correspond to the same set of parameters used for the ELECTRA-base model, as described in~\citep{Clark20}.
For tokenization we instead resorted to the WordPiece scheme adopted by BERT.

It is well known that generating models which adhere to the pre-training and fine-tuning paradigm is a highly resource-intensive process. 
In the case of BERT, for example, the computational complexity of the self-attention layer increases quadratically with the length of the sequences after tokenization.
As a result, limiting the maximum length of the input sequence to 128, at least for the first part of the training, and then increasing it to 512 at a later stage, is a commonly adopted strategy to gain efficiency.
This is done despite the risk of reducing the ability of the model to capture long-distance dependencies and, therefore, to negatively affect its performance.
Differently from other approaches, by leveraging the computational advantages provided by the ELECTRA framework, we were able to investigate the use of a maximum sequence length of 512, instead of 128, for the whole pre-training phase, while keeping the training time on par with other methods.
Indeed, pre-training ELECTRAMed took $\sim$10 days by using one TPU v3 with 8 cores.

\begin{table}[!t] \centering
\caption{Hyperparameters used for ELECTRAMed pre-training}
\label{pretraining_hp}
{
\begin{tabular}
{
>{\raggedright\arraybackslash}p{4cm}
>{\centering\arraybackslash}p{1cm}
}
\toprule 
 \textbf{Hyperparameter} & \textbf{Value} \\ \midrule
 Number of layers & 12  \\
 Hidden size & 768  \\
 FFN inner hidden size & 3072  \\
 Attention heads & 12 \\
 Attention head size & 64 \\
 Embedding Size & 768  \\
 Generator Size & 1/3  \\
 Mask percent & 15 \\
 Learning Rate Decay & Linear \\
 Warmup steps & 10000 \\
 Learning Rate & 2e-4  \\
 Adam $\epsilon$ & 1e-6 \\
 Adam $\beta_1$ & 0.9 \\
 Adam $\beta_2$ & 0.999 \\
 Attention Dropout & 0.1 \\
 Dropout & 0.1\\
 Weight Decay & 0.01 \\
 Batch size & 256  \\
 Train epochs & 1M  \\ \midrule
\end{tabular}}{}
\end{table}

\subsection{ELECTRAMed fine-tuning}
After pre-training, ELECTRAMed was fine-tuned and tested on three biomedical NLP tasks, represented by named entity recognition (NER), relationship extraction (RE) and question answering (QA).

Named entity recognition (NER) is aimed at automatically finding and tagging in a text meaningful terms, called named entities.
In a general domain these typically refer to sequences of words corresponding to specific entities in the real world, such as locations, persons, organizations.
In the biomedical field named entities can represent the name of genes, diseases, chemical compounds, and drugs, to name a few.
The goal of NER is tagging each token in a sentence with one taken from a list of possible named entities, or with the "no entity" label.
NER is often applied as a preliminary step for many applications, such as relationship extraction and knowledge base completion.
It is worthwhile to notice that, compared to the general domain, NER in the biomedical field is considered to be more complex, since biomedical entities constantly grow in number with the scientific progress, may contain special characters and can be referred to using a wide variety of synonyms and abbreviations.

Among the available annotation schemes used to label multi-token named entities, in the present work we resorted to BIO tagging \citep{Sang00}, for which three binary classifiers are trained to label each token in the text as \emph{B} (the token is the beginning of a named entity), \emph{I} (the token is inside a named entity but is not the beginning), and \emph{O} (otherwise).

For fine-tuning and testing ELECTRAMed on biomedical NER, three publicly available corpora were used. 
The first, denoted here as NCBI-disease, is the disease corpus of the National Center for Biotechnology Information (NCBI) \citep{Dogan14}, which is a collection of 793 PubMed abstracts annotated at the mention and concept level.
In detail, it contains 6892 disease mentions mapped into 790 unique disease concepts.
The second dataset (BC5CDR) was used for the BioCreative V Chemical Disease Relation task and is composed by 1500 PubMed articles with 4409 annotated chemicals, 5818 diseases and 3116 chemical-disease interactions \citep{Li2016}.
From the entire collection we extracted the subset with chemical and disease entities, along the lines of previous studies.  
The third dataset (JNLPBA) was released for an open challenge and derives from the GENIA corpus, which is a collection of abstracts retrieved by controlled search on MEDLINE.
For the shared task, the authors simplified the original 36 classes into five super-classes represented, respectively, by proteins, DNA, RNA, cell types and cell lines \citep{Kim04}.
These correspond to the biological named entities of interest.

\begin{table}[!t]\centering
\caption{Description of the corpora used as benchmarks for biomedical NER\label{ner_corpora_table}}
{
\begin{tabular}
{
>{\raggedright\arraybackslash}p{2.5cm}
>{\raggedright\arraybackslash}p{2.5cm}
>{\raggedright\arraybackslash}p{2.5cm}
>{\centering\arraybackslash}p{1.5cm}
>{\centering\arraybackslash}p{1.5cm}
>{\centering\arraybackslash}p{1.5cm}
}
\toprule
\textbf{Dataset} & \textbf{Entity Type} & \textbf{N. entities} & \textbf{N. train} & \textbf{N. val} & \textbf{N. test} \\ \midrule
 NCBI-disease & Disease & 6,892 dis. & 5,429 & 923 & 941\\ \midrule
 BC5CDR & Disease and & 10,227 TOT & 3,951 & 3,957 & 4,145\\
  & Chemical & 4,409 chemicals \\
  &          & 5,818 diseases \\ \midrule
 JNLPBA & Gene and & 59,963 TOT & 16,845 & 1,743 & 3,869\\
  & Protein & 35,336 proteins  \\ 
  &         & 10,589 DNA  \\ 
  &         &  1,069 RNA  \\ 
  &         &  8,639 cell types  \\ 
  &         &  4,330 cell lines  \\ \midrule
\end{tabular}}{}
\end{table}

Relationship extraction (RE) generally follows NER and is aimed at finding semantic relationships which may occur in a text between two or more entities.
In the biomedical domain relationships are extracted between biological entities and, for example, can take the form of different kinds of relations among genes and diseases or of protein-protein interactions.
RE is cast as a text classification problem in which, given a pair of entities, the goal is to assign the correct type of relationship, whether this exists. 
To validate ELECTRAMed on RE problems two datasets were analyzed.
The first is CHEMPROT \citep{Krallinger17}, which is a corpus containing annotated relationship types between chemicals and proteins. From the initial group of 10 different kinds of relationships, five of these (CPR:3, CPR:4, CPR:5, CPR:6 and CPR:9) were used for evaluation.
The second corpus (DDI-2013) contains 792 texts extracted from the DrugBank database and other 233 MEDLINE abstracts describing drug-drug interactions (DDIs).
The corpus was annotated by considering four different types of interactions (i.e. effect, mechanism, advice, and int), where the last was used when a DDI appeared in the text without providing any specific additional information \citep{Herrero13}.

\begin{table}[!t] \centering
\caption{Description of the corpora used as benchmarks for biomedical RE} \label{re_corpora_table} 
{
\begin{tabular} 
{
>{\raggedright\arraybackslash}p{2.4cm}
>{\raggedright\arraybackslash}p{2.5cm}
>{\raggedright\arraybackslash}p{2.5cm}
>{\centering\arraybackslash}p{1.3cm}
>{\centering\arraybackslash}p{1.3cm}
>{\centering\arraybackslash}p{1.3cm}
}
\toprule
\textbf{Dataset} & \textbf{Relationship} & \textbf{N. relations} & \textbf{N. train} & \textbf{N. val} & \textbf{N. test}\\ \midrule
 CHEMPROT & Chemical-  & 10,028 TOT & 19,460 & 11,820 & 16,943 \\ 
  & Protein & 1,983 CPR:3  \\ 
  &         & 5,006 CPR:4  \\ 
  &         & 484 CPR:5  \\ 
  &         & 727 CPR:6  \\ 
  &         & 1,828 CPR:9  \\ \midrule
 DDI-2013 & Drug-Drug & 2,937 TOT & 18,779 & 7,244 & 5,761 \\ 
  & & 1,212 effect  \\ 
  & & 946 mechanism  \\ 
  & & 633 advice  \\ 
  & & 146 int  \\ \midrule
\end{tabular}}{}
\end{table}

Question answering (QA) can be considered an extension of information retrieval and has the purpose of delivering direct and precise responses to questions asked in natural language. 
Focusing on the biomedical domain, and in particular on inquiries about the treatment of a disease, a QA tool would provide, as answers, specific drugs which are effective against that disease or short text passages containing the response.
Indeed, there are three categories of questions which are typically asked to a QA system: confirmation questions, that can be dealt with by yes-or-no statements, list questions, which expect a list of entities or facts, and factoid questions, which require a single short phrase or sentence as response.
In our work, we focused on this last type of questions, for which the goal is finding the correct answer inside a passage.
This is achieved by training a model capable of predicting the starting and the ending tokens of the text segment containing the full expected response.
For biomedical semantic QA we selected as benchmark the pre-processed version of the BioASQ Task 7b-factoid dataset \citep{Yoon20}, which contains both biomedical questions and gold standard answers, in the form of relevant concepts, articles, snippets, exact answers, summaries and the corresponding full PubMed abstracts as passages.

As for NER, question answering in the biological domain poses a major challenge in contrast to the open or other restricted domains, for the presence of a highly specialized terminology and for a potential larger gap in technicality between the questions made by non-expert users and the target documents.
Another critical issue is the scarcity of labelled datasets.
To tackle this problem, prior to addressing the biomedical QA task, we fine-tuned ELECTRAMed on the Stanford Question Answering Dataset (SQuAD v1.1), a large-scale general-domain reading comprehension dataset published by \citep{Rajpurkar16} and including 87,599 and 10,570 examples for training and testing, respectively.
This additional fine-tuning was performed in line with previous studies, which showed its effectiveness in improving the performance on domain-specific activities \citep{Wiese2017}, \citep{Yoon20}.

\begin{table}[!t] \centering
\caption{Description of the corpus (BioASQ 7b-factoid) used as benchmark for biomedical QA}
\label{qa_corpus_table}
{
\begin{tabular}
{
>{\raggedright\arraybackslash}p{2cm}
>{\centering\arraybackslash}p{3cm}
>{\centering\arraybackslash}p{4cm}
}
\toprule 
\textbf{Batch} &
\textbf{N. questions} & \textbf{N. question-context pairs}\\ \midrule
 Train  & 556 & 5537 \\ \midrule
 Test batch 1 & 39 & 98 \\ 
 Test batch 2 & 25 & 56 \\ 
 Test batch 3 & 29 & 84 \\ 
 Test batch 4 & 34 & 90 \\ 
 Test batch 5 & 35 & 79 \\ \midrule
\end{tabular}}{}
\end{table}

The description of the corpora used for the three NLP tasks in terms of number of training, testing and validation examples, and number of observations for each class, are provided in Tables~\ref{ner_corpora_table},~\ref{re_corpora_table} and~\ref{qa_corpus_table}.
The hyperparameters applied for fine-tuning ELECTRAMed on all the tasks are instead indicated in Table~\ref{finetuning_hp}.
Notice that, the datasets at hand are gold standards for the evaluation of biomedical NLP tools.
To enable a fair comparison with previous studies, therefore, we adopted the same partition into training, test and validation sets as originally proposed by the authors of the corpora.
Finally, by way of example, Table~\ref{table_samples} shows the detected named entities and the answers provided by ELECTRAMed for samples extracted from the corpora used for NER and QA tasks.

\begin{table}[!t] \centering
\caption{Hyperparameters used for ELECTRAMed fine-tuning}
\label{finetuning_hp}
{\begin{tabular}
{
>{\raggedright\arraybackslash}p{4cm}
>{\centering\arraybackslash}p{1.5cm}
>{\centering\arraybackslash}p{1.5cm}
>{\centering\arraybackslash}p{2cm}
>{\centering\arraybackslash}p{2cm}
}
\toprule 
\textbf{Hyperparameter} & \textbf{NER} & \textbf{RE} & \textbf{QA\_SQuAD} & \textbf{QA\_BioASQ} \\\midrule
 Learning Rate & 5e-5 & 5e-5 & 3e-5 & 5e-6 \\
 Adam $\epsilon$ & 1e-6 & 1e-6 & 1e-6 & 1e-6 \\
 Adam $\beta_1$ & 0.9 & 0.9 & 0.9 & 0.9 \\
 Adam $\beta_2$ & 0.999 & 0.999 & 0.999 & 0.999 \\
 Layerwise LR Decay & 0.8 & 0.8 & 0.8 & 0.8 \\
 Learning Rate Decay & Linear & Linear & Linear & Linear \\
 Attention Dropout & 0.1 & 0.1 & 0.1 & 0.1 \\
 Dropout & 0.1 & 0.1 & 0.1 & 0.1 \\
 Weight Decay & 0 & 0 & 0 & 0 \\
 Batch size & 32 & 32 & 16 & 16 \\
 Max Sequence Length & 128 & 128 & 384 & 384 \\
 Document Stride & NA & NA & 128 & 128 \\ \midrule 
\end{tabular}}{}
\label{tbl_finetune_hp}
\end{table}

\begin{table*}[!t]
\caption{Detected named entities and answers provided by ELECTRAMed (in bold) for samples taken from NER and QA corpora}
\label{table_samples}
\centering 
{\begin{tabular} {p{2cm}p{2cm}p{11cm}} 
\hline \\[-6pt]
\textbf{Task} & \textbf{Dataset} & \textbf{Sample} 
\\ [+4pt]
\hline 
\\[-6pt]
NER & BC5CDR & Thus, \textbf{indomethacin} by inhibition of \textbf{prostaglandin} synthesis may diminish the blood pressure maintaining effect of the stimulated effect of the stimulated renin-\textbf{angiotensin} system in \textbf{sodium} and volume depletion. \\ 
    &   &   \\
NER & NCBI-disease & Occasional missense mutations in ATM were also found in \textbf{tumour} DNA from patients with \textbf{B-cell non-Hodgkins lymphomas} (\textbf{B-NHL}) and a \textbf{B-NHL} cell line.. \\ 
\\[-6pt]
\hline
\\[-6pt]

QA & BioASQ-7b & Q: What is the cause of a STAG3 truncating variant? \\ 
  &           & A: STAG3 truncating variant as the cause of \textbf{primary ovarian insufficiency}. \textbf{Primary ovarian insufficiency} (POI) is a distressing cause of infertility in young women ... \\
    &   &   \\ 
QA & BioASQ-7b & Q: Which receptor is targeted by Erenumab? \\ 
  &           & A: ... Erenumab, a human monoclonal antibody that inhibits the \textbf{calcitonin gene-related peptide receptor}, is being evaluated for migraine prevention ... \\ 
\\[-6pt]  
\hline
\end{tabular}}{}
\end{table*}

\section{Results}
For named entity recognition the performance of ELECTRAMed was evaluated by means of the F$_1$-score (F), defined as the harmonic mean between precision (P) and recall (R).
Accordingly to \citep{Sang00}, precision was computed as the
percentage of named entities found by the model that were correct (i.e. were an exact match of the corresponding entity in the corpus), whereas recall was set as the percentage of entities included in the corpus and detected by the model.
For the purpose of relationship extraction, the F$_1$-score was applied as well.
In this case, the predicted triplets entity-relation-entity were deemed correct if the relation and the two entities were the same as the ground truth.
Finally, for question answering we resorted to three quality measures, commonly used for assessing tools which generate a list of possible responses to a given inquiry. 
The first is the mean reciprocal rank (MRR), given by the average of the reciprocal ranks of the results for a sample of queries.
The other two are the strict accuracy (SACC) and the lenient accuracy (LACC), for which a question is correctly answered if the gold response is the first element of the list returned by the model, or it is included in the list, respectively.

For each of the NLP problems ELECTRAMed was compared with the current best state-of-the-art (SOTA) models, including those developed by the participants of the 7th BioASQ Challenge runs (QA activity).
The models comprised within each SOTA group are specified in the captions of Tables~\ref{table_ner_results},~\ref{table_re_results} and~\ref{table_qa_results}, for every task and corpus.
These tables also contain missing values denoted as ``?'', to indicate the unavailability of the corresponding measure for the given pair model-dataset.
Missing outcomes are associated to precision and recall for both NER and RE tasks.
For the sake of completeness, we deemed relevant to provide the results for ELECTRAMed also in terms of these two quality measures.
It is also worthwhile to observe that, all the values referred to ELECTRAMed in the tables are obtained by averaging the respective metrics over five runs with different seeds.
This is in line with the experimental settings adopted for SciBERT \citep{Beltagy19}, for which the outcomes were computed as average over multiple runs. 
For the other SOTA models, instead, we were unable to establish whether the results reported by the authors were averaged or corresponded to the best outcomes achieved.

The results obtained by ELECTRAMed for the task of named entity recognition are shown in Table~\ref{table_ner_results}.
The proposed model reached the highest F$_1$-score (90.03) on the BC5CDR dataset.
By performing better than the current SOTA approaches, ELECTRAMed sets a novel state-of-the-art performance on this corpus for NER purposes in terms of F$_1$-score.
For NCBI-disease and JNLPBA datasets, ELECTRAMed was not among the top three SOTA models, but still reached  comparable results on the first corpus.

\begin{table}[!t] \centering
\caption{Precision (P), recall (R) and F$_1$-score (F) for the NER task}
\label{table_ner_results}
{
\begin{tabular}
{
>{\raggedright\arraybackslash}p{2cm}
>{\centering\arraybackslash}p{1.5cm}
>{\centering\arraybackslash}p{1.5cm}
>{\centering\arraybackslash}p{1.5cm}
>{\centering\arraybackslash}p{1.5cm}
>{\centering\arraybackslash}p{3cm}
}
\toprule
\textbf{Benchmark} & \textbf{Metrics} & \textbf{SOTA1} & \textbf{SOTA2} & \textbf{SOTA3} & \textbf{ELECTRAMed}\\ \midrule
NCBI-disease & P & 88.22 & ? & ? & 85.87 \\
             & R & 91.25 & ? & ? & 89.29 \\
             & F & \textbf{\underline{89.71}} & 89.13 & 88.85 & 87.54 \\
BC5CDR       & P & 92.05 & ? & ? & 88.76 \\
             & R & 87.91 & ? & ? & 91.34 \\
             & F & 89.93 & 89.73 & 89.42 & \textbf{\underline{90.03}} \\
JNLPBA       & P & ? & 72.24 & ? & 69.33 \\
             & R & ? & 83.26 & ? & 78.56 \\
             & F & \textbf{\underline{81.29}} & 77.59 & 77.03 & 73.65 \\ \midrule
\end{tabular}
}
\\
{State-of-the-art (SOTA) performance. For NCBI-disease, SOTA1 is BioBERT, \citep{Lee2020}, SOTA2 is Spark NLP, \citep{Kocaman20}, SOTA3 is BioFLAIR, \citep{Sharma19}. For BC5CDR, SOTA1 is RL+DS+PA, \citep{Nooralahzadeh2019}, SOTA2 is Spark NLP and SOTA3 is BioFLAIR. For JNLPBA, SOTA1 is Spark NLP, SOTA2 is BioBERT and SOTA3 is BioFLAIR.}
\end{table}

The outcomes for the task of relationship extraction are indicated in Table~\ref{table_re_results}.
These results clearly demonstrate the superior effectiveness of SciBERT over the other models, but also show the promising performance of ELECTRAMed.
Indeed, on the DDI-2013 dataset the proposed approach provided results which are close to the second-best model (SOTA2) and ranked at the third position by reaching a fairly higher F$_1$-score compared to the current SOTA3 (79.13 vs. 72.90).

Finally, the results achieved for the question answering task are depicted in Table~\ref{table_qa_results}.
ELECTRAMed was able to outperform all the competitors in two (batches 1 and 4) out of the five runs, and to provide comparable results with the best approach (KU-DMIS-5) for the second batch.
For the remaining runs (batches 3 and 5), ELECTRAMed ranked at the seventh and sixth position, respectively, among all the participants.
To investigate the ability of ELECTRAMed of providing high-quality responses across the whole challenge, besides the BioASQ baseline method we selected the models that competed in all the runs and, for each of them, we computed a score given by the ratio between their MRR and the highest MRR reached in a given batch.
For each model, the sum of the scores over all the runs can be seen as a measure of its ability of providing responses which are close, if not equal, to the best ones in terms of MRR.
The results of this analysis are reported in Table~\ref{table_qa_score} and support the effectiveness of ELECTRAMed for the QA task at hand.
The proposed model, indeed, achieved the highest total score being associated with ratios more densely distributed around 1. 
In particular, ELECTRAMed performed better than systems based on the BioBERT approach (KU-DMIS-5 model).

\begin{table}[!t] \centering
\caption{Precision (P), recall (R) and F$_1$-score (F) for the RE task} 
\label{table_re_results}
{\begin{tabular} 
{
>{\raggedright\arraybackslash}p{2cm}
>{\centering\arraybackslash}p{1cm}
>{\centering\arraybackslash}p{1.5cm}
>{\centering\arraybackslash}p{1.5cm}
>{\centering\arraybackslash}p{1.5cm}
>{\centering\arraybackslash}p{2.5cm}
}
\toprule
\textbf{Benchmark} & \textbf{Metrics} & \textbf{SOTA1} & \textbf{SOTA2} & \textbf{SOTA3} & \textbf{ELECTRAMed}\\\midrule
CHEMPROT     & P & ? & 77.02 & ? & 75.47 \\
             & R & ? & 75.90 & ? & 70.67 \\
             & F & \textbf{\underline{83.64}} & 76.46 & 74.40 & 72.94 \\
DDI-2013      & P & ? & ? & 74.10 & 80.07 \\
             & R & ? & ? & 71.80 & 78.24 \\
             & F & \textbf{\underline{84.08}} & 79.90 & 72.90 & 79.13 \\ \midrule
\end{tabular}}
\\
{State-of-the-art (SOTA) performance. For CHEMPROT, SOTA1 is SciBERT, \citep{Beltagy19}, SOTA2 is BioBERT, SOTA3 is BlueBERT, \citep{Peng19}. For DDI-2013, SOTA1 is DESC+MOL+SciBERT, \citep{Asada20}, SOTA2 is  BlueBERT, SOTA3 is Hierarchy Bi-LSTMs +Att.+SDP, \citep{Zhang18}.}
\end{table}

\begin{table}[!h] \centering
\caption{Strict accuracy (SACC), lenient accuracy (LACC) and mean reciprocal rank (MRR) for the QA task}
\label{table_qa_results}
{\begin{tabular}
{
>{\centering\arraybackslash}p{1cm}
>{\raggedright\arraybackslash}p{3cm}
>{\centering\arraybackslash}p{1.5cm}
>{\centering\arraybackslash}p{1.5cm}
>{\centering\arraybackslash}p{1.5cm}
}
\toprule 
\textbf{Batch} & \textbf{Competitor} & \textbf{SACC} & \textbf{LACC} & \textbf{MRR} \\\midrule
1 & (1) ELECTRAMed & 44.62 &	51.28 &	\textbf{\underline{47.95}} \\
 & (2) KU-DMIS-1 & 41.03 &	53.85 &	46.37 \\
 & (3) BJUTNLPGroup & 30.77 &	41.03 &	34.83 \\
 & (4) auth-qa-1 & 25.64 &	30.77 &	27.78 \\ \midrule
2 & (1) KU-DMIS-5 & 52.00 &	64.00 &	\textbf{\underline{56.67}} \\
 &  (2) ELECTRAMed & 46.40 &	62.40 &	53.16 \\
 & (3) QA1 & 36.00 &	48.00 &	40.33 \\
 & (4) transfer-learning & 24.00 &	44.00 &	32.67 \\ \midrule
3 & (1) QA1 & 44.83 &	58.62 &	\textbf{\underline{51.15}} \\
 & (2) UNCC\_QA\_1 & 44.83 &	58.62 &	51.15 \\
 & (3) google-gold-input & 41.38 &	65.52 &	50.23 \\
 & (7) ELECTRAMed & 37.93 &	58.62 &	46.62 \\ \midrule
4 & (1) ELECTRAMed & 61.18 &	82.35 &	\textbf{\underline{69.55}} \\
 & (2) KU-DMIS-1 & 58.82 &	82.35 &	69.12 \\
 & (3) FACTOIDS & 52.94 &	73.53 &	61.03 \\
 & (4) UNCC\_QA3 & 52.94 &	73.53 &	61.03 \\ \midrule
5 & (1) KU-DMIS-5 & 28.57 &	51.43 &	\textbf{\underline{36.38}} \\
 & (2) BJUTNLPGroup & 28.57 &	40.00 &	33.81 \\
 & (3) UNCC\_QA\_1 & 28.57 &	42.86 &	33.05 \\ 
 & (6) ELECTRAMed & 24.57 &	44.00 &	31.42 \\ \midrule
\end{tabular}}
\\
{\textit{Note}: The number in round brackets beside each model indicates the ranking in the challenge run.}
\end{table}

\begin{table}[!h] \centering
\caption{Scores over the five runs of the 7h BioASQ-factoid Challange} 
\label{table_qa_score}
{\begin{tabular}
{
>{\raggedright\arraybackslash}p{3cm}
>{\centering\arraybackslash}p{1cm}
>{\centering\arraybackslash}p{1cm}
>{\centering\arraybackslash}p{1cm}
>{\centering\arraybackslash}p{1cm}
>{\centering\arraybackslash}p{1cm}
>{\centering\arraybackslash}p{1cm}
}
\toprule \textbf{Competitor} & \textbf{Batch1} & \textbf{Batch2} & \textbf{Batch3} & \textbf{Batch4} & \textbf{Batch5} & \textbf{Total}\\\midrule
BioASQ Baseline & 0.323 & 0.241 & 0.258 & 0.364 & 0.238 & 1.424\\
auth-qa-1       & 0.579 & 0.541 & 0.669 & 0.536 & 0.412 & 2.737\\
KU-DMIS-1       & 0.967 & 0.771 & 0.924 & 0.994 & 0.886 & 4.542\\
LabZhu,FDU      & 0.120 & 0.441 & 0.775 & 0.699 & 0.615 & 2.650\\
ELECTRAMed      & \textbf{\underline{1.000}} & 0.938 & 0.911 & \textbf{\underline{1.000}} & 0.864 & \textbf{\underline{4.713}}\\
            \midrule
\end{tabular}}
\\
{\textit{Note}: Three models by Lab Zhu at Fudan University were proposed. The table includes the one that performed better across the runs.}
\end{table}

%
%

\section{Conclusions \& Future Developments}
The recent literature on biomedical NLP has been heavily influenced by BERT-based architectures and by the usage of domain-specific corpora for pre-training.
Meanwhile, in the general-domain NLP literature, a plethora of transformer-based architectures have flourished, bringing significant improvements to the first wide-spread implementation.
In this study we presented ELECTRAMed, a new pre-trained language model for biomedical NLP.
ELECTRAMed was pre-trained from scratch on a biomedical corpus using a domain-specific vocabulary, and was shown to obtain valuable results for some of the most commonly addressed NLP tasks arising in the biomedical field.
For ELECTRAMed pre-training we leveraged the efficiency of the ELECTRA architecture, and we were able to perform at par of the current state-of-the-art models while keeping  the computational effort low, in terms of both time and cost.

The results achieved in the present work encourage future studies that can be undertaken along different directions.
From one side, it would be worthwhile to investigate the performance of ELECTRAMed when the maximum sequence length of the input is reduced from 512 to 128 for the entire, or at least, the most part of the training phase, with the aim of further reducing the computational requirements.
From the other, it would be useful to explore the impact on the model performance of using a vocabulary built upon the selected pre-training corpus.
With this study we hope to spark a new wave of transformer-based architectures in the biomedical domain.
Consequently, as a future research line it would be also interesting to investigate potential improvements for tasks related to biomedical information extraction by combining existing biomedical domain knowledge resources (e.g. knowledge bases) into novel transformer-based learning frameworks.

\section*{Acknowledgements}

We thank Dr. Sandra Coecke from the Joint Research Center at European Commission and Dr. Anna Beronius from Karolinska Institute for their valuable and fruitful discussions that fostered a positive and encouraging environment which greatly contributed to the development of our work.

\bibliographystyle{unsrt}  
\bibliography{references}  


\end{document}